# Predicting Individual Differences in Lipreading From Eye-Tracking and Pupillometry During Silent Visual Speech

Puneet Kumar[1,4], Satyarth Malik[2], Gao Zheng[3], Guoying Zhao[4], and Xiaobai Li[3,4]

[1]Indian Institute of Technology Ropar, Rupnagar, Punjab, India

[2]Indian Institute of Technology (BHU) Varanasi, Uttar Pradesh, India

[3]Zhejiang University, Hangzhou, China

[4]University of Oulu, Oulu, Finland

Puneet Kumar and Satyarth Malik contributed equally to this work.

Correspondence: Xiaobai Li, [3]Zhejiang University, Hangzhou, China. E-mail: xiaobai.li@zju.edu.cn

Manuscript type: Research Article

Approximate word count: 11,000 (abstract and main text, excluding references)

Number of figures: 5   Number of tables: 2

## Abstract

**Purpose:** Lipreading ability varies widely, including among people who are hard of hearing. Although mouth-directed attention is linked to speechreading outcomes, it remains unclear how much lipreading skill can be inferred from viewing behavior and which signals carry this information. We examine whether individual differences in lipreading ability are reflected in viewing behavior during silent speech and quantify the contributions of gaze dynamics, pupil-linked signals, and spatial viewing allocation.

**Method:** We introduce the Lipreading Perception Study (LiPS) dataset and a tracker-independent preprocessing pipeline that synchronizes SR Research EyeLink gaze and pupil recordings with the stimuli. Twenty-four adults (age 32 ± 10 years, mean ± SD) completed a four-alternative forced-choice silent lipreading task. An extreme-groups protocol yielded 18 participants labeled as stronger or weaker lipreaders. Each participant–trial included four aligned views: lip-region images, frame-level gaze, pupil summaries, and spatial viewing proportions. A four-branch neural network with cross-modal attention predicted labels using participant-disjoint splits and systematic ablations.

**Results:** The full model reached accuracy 0.680 ± 0.040 (mean ± SD across five folds), outperforming naive fusion baselines in accuracy. The pupil and viewing-summary branch was most informative: removing it caused the largest degradation (0.618 accuracy, 0.623 AUC), while it alone achieved 0.675 accuracy. The lip-image stream contributed little to participant-level discrimination.

**Conclusions:** Individual differences in lipreading are reflected in pupil-linked and frame-level gaze signals rather than in how much time a viewer spends looking at the mouth. Synchronized gaze and pupillary measures may provide objective markers of visual speech perception, warranting investigation in larger, clinically diverse populations. LiPS provides a resource for extending this work.

**Keywords:** visual speech perception, lipreading, speechreading, eye tracking, gaze, pupillometry, individual differences, processing effort

## 1. Introduction

Humans can recover a surprising amount of linguistic information from a talker's facial motion alone. This ability, known as lipreading or speechreading, supports communication in noisy environments and especially for people who are deaf or hard of hearing, for whom the visible face is a primary source of spoken-language information (Bernstein, Jordan, et al., 2022; Lalonde & McCreery, 2020; Santana et al., 2025). Despite its importance, lipreading ability varies considerably across individuals, and the mechanisms underlying this variation remain only partly understood (Battista et al., 2025). Existing reviews identify three major challenges: how lipreading skill should be measured, what factors drive it, and how it can be effectively trained (Battista et al., 2025; Bernstein, Jordan, et al., 2022).

Three lines of evidence motivate the study. First, individual differences in lipreading are substantial and stable: audiovisual benefit is predictable from unimodal speech perception (von Seth et al., 2025), changes across the lifespan (Tye-Murray et al., 2016), and is associated with stable face-viewing preferences (Viktorsson et al., 2023). Lipreading is also malleable, improving with structured feedback (Bernstein, Auer, & Eberhardt, 2022) and targeted training in children (Buchanan-Worster et al., 2021). Second, eye gaze provides a window into visual speech perception: viewers shift toward the mouth as listening conditions worsen (Yi et al., 2013), although this relationship varies across tasks (Wilson et al., 2016). Recent work shows that gaze during silent speech tracks the unheard speech envelope (Benz et al., 2025) and relates to cortical tracking of audiovisual speech (Tan et al., 2022), suggesting temporal gaze dynamics carry more information than static dwell-time summaries. Third, pupillometry provides a complementary measure of processing effort (Zekveld et al., 2018), is sensitive to task demands (Colby & McMurray, 2021; Keur-Huizinga et al., 2024; Kuchinsky et al., 2023), and has utility in classifying listener hearing status (Lebiecka-Johansen et al., 2025).

Two practical barriers stand in the way of studying these differences: datasets that jointly contain silent lipreading stimuli, behavioral responses, frame-synchronized gaze, and pupillary signals are scarce, and reproducible pipelines that align heterogeneous eye-tracking recordings with stimulus video at frame resolution are largely unavailable. We introduce the Lipreading Perception Study (LiPS) dataset and a schema-unified preprocessing pipeline addressing both, following recommendations for transparent eye-tracking operationalization (Hessels et al., 2018; Hooge et al. 2025).

Using this resource, we formulate lipreading assessment as predicting a viewer's lipreading ability from their viewing behavior during silent speech. Unlike machine lipreading, which determines *what* the talker said (Afouras et al., 2018; Assael et al., 2016; Chung & Zisserman, 2016; Sheng et al., 2024), our objective is to estimate *how well* a human observer understands the talker. Each participant–trial is represented through four complementary views: lip-region image sequences, frame-level gaze dynamics, pupil-derived physiological descriptors, and coarse spatial gaze distributions, modeled using branch-specific encoders with attention-based fusion (Altinok, 2024; Girdhar et al., 2023; Jiang et al., 2024; Zhong et al., 2024).

### 1.1 Visual Speech Perception, Gaze, and Individual Differences

During audiovisual speech perception, attention shifts toward the mouth as listening conditions worsen (Yi et al., 2013), and increased mouth fixation has been associated with improved speechreading (Worster et al., 2018), though this relationship depends on the task, stimulus, and population (Wilson et al., 2016). More recent work demonstrates adaptive gaze strategies during degraded speech (Banks et al., 2021; Slomianka et al., 2025), suggesting that static spatial summaries capture only part of the information in viewing behavior.

Recent JSLHR research has examined eye gaze as a behavioral component of visual and audiovisual speech perception. Eye-tracking studies show that visual attention strategies vary with sensory experience and language familiarity, with adults who are deaf directing gaze primarily to the mouth and adults with typical hearing primarily to the eyes (Wang et al., 2020). Under degraded audiovisual speech, longer fixations on the talker's mouth have been associated with better overall recognition accuracy, although visual cues benefited perceptual adaptation less than previously assumed (Banks et al., 2021). Notably, that association involved fixation duration rather than total mouth-directed viewing time, and in visual-only paradigms with adults the proportion of gaze time directed at the mouth has not predicted speechreading accuracy (Lansing & McConkie, 2003; Wilson et al., 2016), although it did in children (Worster et al., 2018). More recent work further indicates that the contribution of visual speech to audiovisual recognition is smaller than earlier estimates and varies by viseme category (O'Hanlon et al., 2025). Together, these findings establish gaze as a meaningful behavioral index of visual speech processing while leaving unresolved whether temporally resolved gaze and pupil-linked physiological measures can characterize stable individual differences in visual-only speech perception.

Individual differences in visual speech perception are substantial and stable (Tye-Murray et al., 2016; von Seth et al., 2025), and subject-specific gaze patterns generalize beyond speech tasks (Kollenda et al., 2025; Melnyk et al., 2025). Clinical reviews emphasize persistent individual differences while identifying gaps in measurement and training (Battista et al., 2025; Bernstein, Jordan, et al., 2022). These differences are also modifiable: training improves speechreading in adults (Bernstein, Auer, & Eberhardt, 2022) and children (Buchanan-Worster et al., 2021). Neurophysiological studies further show that visual cortical representations encode linguistic categories predictive of lipreading ability (Nidiffer et al., 2023), auditory cortex contributes selectively during lipreading (Karthik et al., 2024), and lipreading a naturalistic narrative engages neural processes shared with listening and reading (Saalasti et al., 2023).

### 1.2 Pupillometry and Processing Effort

While gaze characterizes the deployment of visual attention, pupillometry provides a complementary measure of cognitive and perceptual effort. Pupil dilation is a well-established objective index of listening effort with demonstrated sensitivity to task demands and arousal (Zekveld et al., 2018). Pupillary responses partly converge with cognitive dual-task measures of effort (Colby & McMurray, 2021), vary systematically with interacting acoustic, lexical, and cognitive demands (Kuchinsky et al., 2023), and are recognized as one of the most sensitive physiological indicators of processing effort within multimodal assessment frameworks (Keur-Huizinga et al., 2024). Pupillometry has demonstrated robust performance across the lifespan, although standardized methodologies remain necessary for broader clinical adoption (Raghavendra, 2025).

Most directly relevant to the present work, pupil-derived features recorded during sentence perception have been successfully used to classify listener hearing status (Lebiecka-Johansen et al., 2025), providing direct precedent for incorporating pupillary signals into predictive models of speech perception. Unlike these predominantly auditory or audiovisual paradigms, the present study focuses on silent visual speech, extending pupillometry to quantify the effort required to recover linguistic information solely from facial movements.

### 1.3 Machine Lipreading and the Distinction From Predicting Human Lipreading

Deep learning has driven rapid progress in machine lipreading and audiovisual speech recognition (AVSR), enabled by large datasets and powerful sequence models (Sheng et al., 2024), with datasets such as OuluVS2 (Anina et al., 2015) and LRW/LRS (Chung & Zisserman, 2016; Chung et al., 2017), and architectures spanning LipNet (Assael et al., 2016), audiovisual fusion (Afouras et al., 2018), masked multimodal learning (Shi et al., 2022), self-supervised visual speech recognition (Narasimharao et al., 2026; Yuan et al., 2025), and synchronized token learning (Ahn et al., 2024).

Recent gaze modeling treats scanpaths as temporally structured signals (Jiang et al., 2024; Nishiyasu & Sato, 2024; Zhong et al., 2024), while multimodal learning supports modality-specific representations and adaptive fusion (Girdhar et al., 2023), and explainable fusion enables interpretation of modality contributions (Altinok, 2024). In human speech perception, research has examined audiovisual integration (O'Hanlon et al., 2025; Wang et al., 2020), individual differences in visual speech processing (Banks et al., 2021), and metacognitive awareness of lipreading benefits (Giovanelli et al., 2025).

However, these studies do not directly predict individual lipreading ability from eye-tracking behavior. Together, this literature informs our architectural design and fusion strategy while distinguishing our prediction target: human lipreading ability rather than talker speech recognition.

### 1.4 Eye-Tracking Resources, Methodology, and the Remaining Gap

Reliable behavioral analysis requires consistent processing and reporting of eye-movement data, as fixation and saccade events are operationalized differently across studies (Hessels et al., 2018), motivating recommendations for transparent eye-tracking methodology (Hooge et al., 2025). The gaze-estimation community has developed datasets spanning laboratory, mobile, and unconstrained settings (Fischer et al., 2018; Kellnhofer et al., 2019; Krafka et al., 2016; Park et al., 2020; Zhang et al., 2015, 2020), complemented by tools such as OpenFace (Baltrusaitis et al., 2016). However, these resources primarily target gaze direction estimation rather than the relationship between viewing behavior and visual speech perception.

As the preceding sections indicate, the components required to study lipreading behavior exist, but no single resource combines them. Predicting viewer-specific lipreading ability requires four synchronized elements per trial: (i) a silent visual speech stimulus of known content; (ii) a behavioral response to quantify lipreading ability; (iii) frame-level gaze aligned with the stimulus; and (iv) synchronous physiological measures of processing effort. Speechreading gaze studies record gaze alongside perception outcomes but report aggregated measures (Banks et al., 2021; Wilson et al., 2016; Worster et al., 2018; Yi et al., 2013); machine lipreading datasets provide visual speech stimuli but no human observer (Anina et al., 2015; Chung & Zisserman, 2016; Chung et al., 2017); gaze-estimation datasets lack a speechreading task (Park et al., 2020; Zhang et al., 2015, 2020); and pupillometry studies focus on auditory or audiovisual rather than silent visual speech (Lebiecka-Johansen et al., 2025; Raghavendra, 2025; Zekveld et al., 2018). Retrospective integration is further hindered by inconsistent event definitions and heterogeneous acquisition protocols (Hessels et al., 2018).

This study integrates all four elements within a common temporal framework, supported by a schema-unified preprocessing pipeline and frame-level synchronization necessary for sequential gaze modeling (Benz et al., 2025). Each participant–trial is represented through complementary temporal, spatial, and physiological views of viewing behavior, which are processed using branch-specific encoders and learned attention-based fusion. Ablation analyses further assess the contribution of individual signals to predictive performance.

### 1.5 The Present Study

This study addresses three questions: (i) Can lipreading ability be predicted from gaze and pupillary behavior during silent speech? (ii) Which behavioral signals contribute most, and does modeling their temporal dynamics improve performance? (iii) What do these signatures reveal about successful lipreading and their potential for assessment and training in people who are hard of hearing? Anticipating the results, the central finding is not that this prediction is possible, but which signals carry the information. Signals from the pupil-and-viewing-summary branch and the frame-level gaze sequence distinguish stronger from weaker lipreaders, whereas the proportion of time spent viewing the mouth does not differ between them. The full attention-fusion model achieves 0.680 ± 0.040 accuracy (AUC 0.700 ± 0.068), exceeding naive fusion baselines in accuracy and matching the best single-branch (pupil/summary-only) model, indicating that fusion adds robustness and interpretability rather than a large gain over that branch alone. Removing the pupil/summary branch causes the largest degradation, while removing the lip-image stream leaves performance essentially unchanged, indicating that pupil-linked and gaze-derived signals drive prediction. This study's major contributions are as follows.

- **A visual-speech-and-gaze dataset.** We introduce a curated dataset linking silent lipreading stimuli, behavioral responses, frame-synchronized gaze trajectories, and pupillary signals for studying individual differences in visual speech perception.
- **A reproducible, tracker-independent preprocessing pipeline.** We develop a schema-unified pipeline that transforms heterogeneous eye-tracker exports (here, EyeLink) into a consistent, event-enriched representation aligned with stimulus video at frame resolution.
- **A behavioral formulation of lipreading assessment.** We formulate lipreading assessment as predicting viewer-specific lipreading ability from synchronized gaze, spatial viewing behavior, and pupillary responses, distinguishing this task from machine lipreading and audiovisual speech recognition.

- **An interpretable multimodal framework.** We propose a four-branch attention-based architecture that integrates heterogeneous behavioral and visual signals through multimodal fusion, enabling both accurate prediction and interpretation of modality contributions.
- **Behavioral analysis of lipreading skill.** We perform ablation and interpretability analyses to identify the gaze- and pupil-linked behavioral signals associated with lipreading skill.

## 2. Method

This section describes the selection of the lipreading assessment and speech material, the collection protocol, task and acquisition setup, preprocessing and frame-level alignment pipeline, the four input views derived from each participant-trial, the model architecture, and the training and evaluation protocol.

### 2.1 Participants

Thirty adults were recruited; six were excluded during recruitment, five because calibration could not be established and one after repeated drift-correction attempts, yielding $N = 24$ participants (aged 32 ± 10 years, mean ± SD) who completed the task (Table 1). Gaze recordings from two of them were unusable, one because of excessive tracking loss and one because calibration accuracy degraded during the session beyond the 0.5° validation criterion, leaving 22 with gaze data; both fell in the mid-range accuracy group and were therefore not part of the classification analyses (Section 2.5). Participants were proficient English speakers from nine countries: India and Finland (6 each), China and Iran (3 each), and Bangladesh (2), Malaysia, Thailand, Brazil, and Ukraine (1 each), with within-group diversity in English exposure. All participants had normal or corrected-to-normal vision, confirmed at recruitment. Hearing status was not formally assessed; because all stimuli were presented without sound (Section 2.2), task performance reflects visual speech perception alone. All participants provided informed consent, and the participant shown in Figure 1 also consented to publication of the image.

## 2.2 Lipreading Assessment and Speech Material

***Lipreading assessment.***

We adopted the Test of Adult Speechreading (TAS) paradigm (Ellis et al., 2001; Mohammed et al., 2003), which spans word, sentence, and story levels in a closed-set format and has an established record of eliciting stable individual differences (Mohammed et al., 2005, 2006). Because original audiovisual materials were unavailable for reuse, we retained TAS's closed-set format but built a new set of silent-video items.

***Speech material.***

We used the Harvard (IEEE) Sentences (IEEE Subcommittee on Subjective Measurements, 1969), a corpus of 720 phonetically balanced sentences, and reused an existing publicly available audiovisual recording contributed by Brown (2020): a young adult female American-English speaker recorded with a laptop webcam. Only the video channel was used; audio was discarded before presentation. Sentences were curated into a graded difficulty structure based on vocabulary commonality, single- versus multi-idea content, and the availability of plausible confusable alternatives, refined through pilot testing. Response choices were built as minimal single-word substitutions across the four options rather than whole-sentence alternatives, since pilot testing showed whole-sentence distractors were resolvable from text alone.

## 2.3 Design Rationale for the Collection Protocol

Each experimental choice directly supports the study objective.

**Silent videos** isolate visual speech perception from hearing threshold and audiovisual integration. **Eye tracking** provides the only direct, continuous measure of visual attention deployment and the temporal resolution needed to align gaze with the stimulus (Viktorsson et al., 2023; Yi et al., 2013). **Pupillometry** complements the "where" and "when" of gaze with "how hard," as pupil dilation is an established index of processing effort (Colby & McMurray, 2021; Lebiecka-Johansen et al., 2025; Zekveld et al., 2018). **Stimulus corpus:** Building the stimulus set from an independent, phonetically balanced corpus with a previously validated recording decouples item difficulty from talker-specific artifacts (Section 2.2). **4AFC** **format** provides objective scoring without transcriber variance, with chance at 0.25; the observed accuracy range (0.25–0.80) confirms adequate discrimination. **Controlled acquisition** keeps language, hardware, display geometry, and illumination constant to minimize confounds, particularly for luminance-sensitive pupil measurements. **Extreme-group labeling** maximizes the contrast between stronger and weaker lipreaders for initial evaluation; graded prediction is left for future work.

### 2.4 Task Overview

The dataset links (i) visual-only lipreading stimuli, (ii) participant responses, and (iii) synchronized, time-resolved gaze and pupil signals. Each participant completes a visual-only lipreading task of 20 short, silent clips of a single talker articulating Harvard sentences (Section 2.2), answering a 4AFC question after each; overall accuracy is the fraction of correctly answered trials, and is our operational definition of lipreading skill. Formally, given a participant $p$ with gaze and viewing data across trials, we predict a binary label $y_p \in \{weaker, stronger\}$ (equivalently, $\{weak, strong\}$) derived from the accuracy thresholds defined below, evaluated at the participant/clip level with participant-disjoint splits (Section 2.9).

### 2.5 Apparatus and Acquisition

Eye movements and pupil size were recorded using an SR Research EyeLink Portable Duo, operating in remote, head-free-to-move mode with a target sticker on the participant's forehead in place of a chin or forehead rest, while participants completed the silent lipreading task on a 1920 × 1080 display at a viewing distance of approximately 60 cm (Figure 1). Both eyes were tracked binocularly at 250 Hz. The eye tracker was placed below the display, centered and facing the participant; a separate tripod-mounted webcam, level with the top of the display, recorded the participant's face as part of the LiPS data collection. Each session opened with a 9-point calibration and validation, repeated until the mean validation error was below 0.5° of visual angle. Stimuli were presented as screen-recorded videos at approximately 30 frames/s under constant, dim room illumination and a fixed clip size and background, so that stimulus luminance was approximately constant across trials, a condition on which the pupil-based effort measures of Section 2.8.3 depend. Data collection was conducted at the University of Oulu, Finland. The preprocessing pipeline is designed to be tracker-independent, and an extension supporting Tobii eye trackers is currently being integrated to enable cross-device evaluation (Section 4.4).

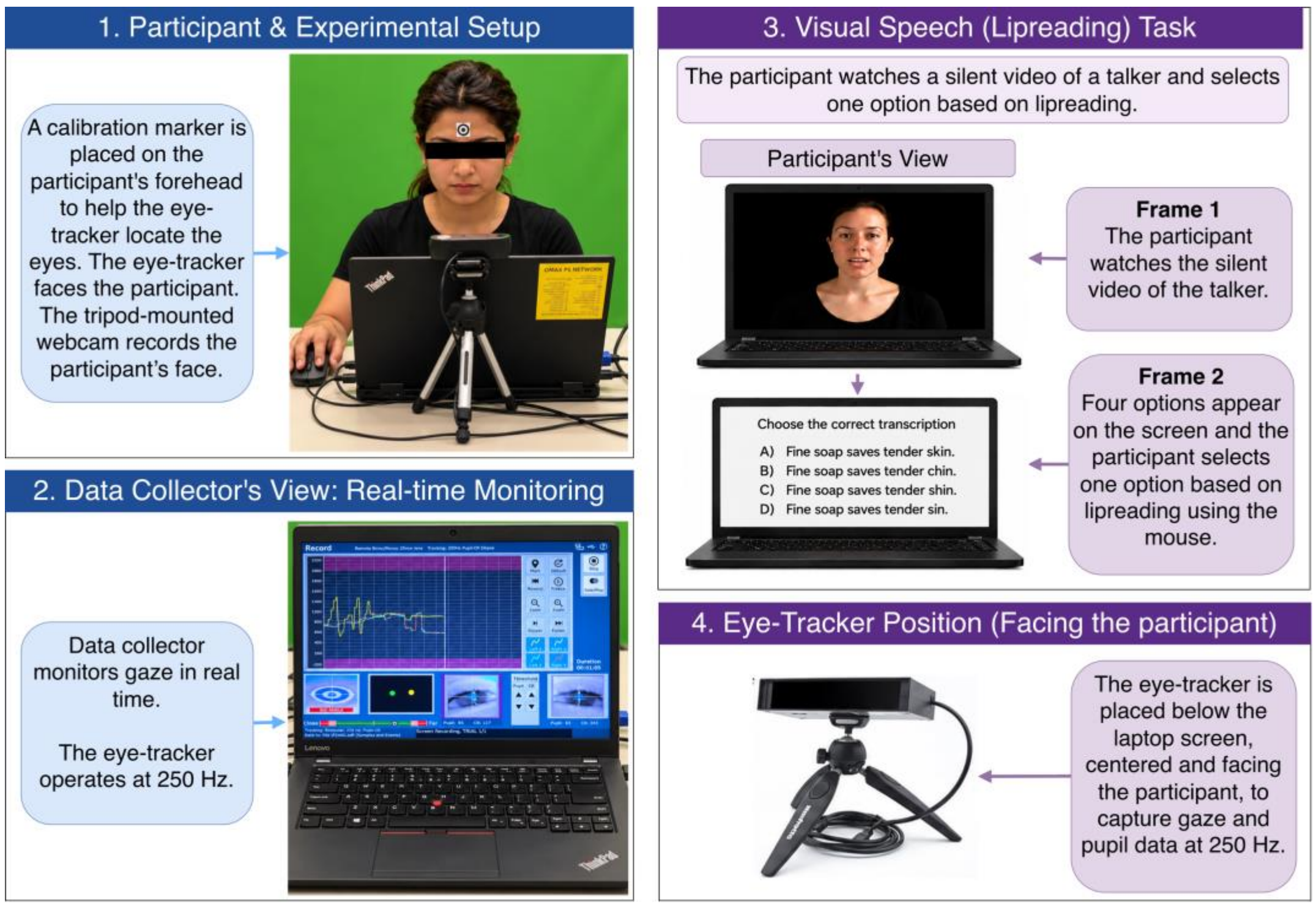

**Figure 1.** Overview of the experimental setup and data collection protocol. The figure illustrates the eye-tracking setup used for recording gaze during silent-video lipreading, including the participant workstation, the stimulus presentation interface with four response options, the eye-tracking device, and the experimenter's live gaze-monitoring interface. Panel 3 shows an illustrative item built from a Harvard sentence with single-word substitutions (Section 2.2). The participant shown consented to publication of this image.

Table 1 summarizes the participants and recordings. Each participant completed 20 silent lipreading trials followed by a four-alternative forced-choice (4AFC) response, from which participant-level lipreading accuracy was computed. For binary classification, participants with accuracy $\geq 0.55$ were labeled as *stronger* lipreaders, those with accuracy $\leq 0.40$ as *weaker* lipreaders, and the six participants between these thresholds were excluded from the classification analyses, yielding an extreme-groups dataset of 8 stronger- and 10 weaker-lipreader participants (Table 1). Session duration is reported separately for participants who completed the task through a locally run presentation script (script) and through a browser-based web interface (web).

**Table 1.** LiPS dataset: participant and recording summary.

| Item | Value |
|---|---|
| Participants (N) | 24 (gaze data available for 22) |
| Age (years, mean ± SD) | 32 ± 10 |
| Gender | 13 M / 11 F |
| Countries represented | 9 (India 6, Finland 6, China 3, Iran 3, Bangladesh 2, Malaysia 1, Thailand 1, Brazil 1, Ukraine 1) |
| Trials per participant | 20 (silent lipreading clips) |
| Response format | 4-alternative forced choice |
| Stimulus frame rate | ≈ 30 fps (screen recording) |
| Stimulus resolution | 1920 × 1080 (screen) |
| Mean session duration (script) | 9.6 min (SD 1.6) |
| Mean session duration (web) | 12.3 min (SD 2.5) |
| Accuracy (mean ± SD) | 0.460 ± 0.140 |
| Accuracy range | [0.25, 0.80] |
| Binary split (stronger/weaker) | 8 / 10 (mid: 6, excluded); 360 segments used for classification |
| Excluded during recruitment | 6 of 30 invited (5 failed calibration entirely; 1 excluded after 6 repeated drift-correction attempts) |

### 2.6 Data Representation, Preprocessing, and Gaze-Video Alignment

The EyeLink system exports multiple reports per recording. For modeling and analysis, we convert each session into a single event-enriched, sample-centric table containing: **identifiers** (participant, trial, stimulus/video ID, and per-trial response/accuracy labels); **timestamps** (eye-tracker timestamps and derived computer time for synchronization); **gaze points** (screen ($x$,$y$) in pixels and normalized coordinates); **pupil and validity** (pupil diameter where available, and validity/blink indicators); and **event channels** (fixation- and saccade-related indices and flags used to derive a compact event-type label).

***Video segmentation.***

Each session contains multiple trials. We identify per-video/per-clip segments using stimulus start/end messages, convert them to a common time base, and assign every gaze sample whose timestamp falls within a segment to the corresponding trial and video ID.

***Frame-level alignment (gaze → video).***

A central requirement, and one on which the sequence models depend, is extracting gaze per video frame. Let the stimulus have frame rate $f \approx 30$ fps. For each gaze sample with segment-relative timestamp $t$ (seconds), we map to a frame index

$$k = \lfloor f \cdot (t + \Delta_p) \rfloor \tag{1}$$

where $\Delta_p$ is a participant-specific offset (seconds) accounting for systematic lag between screen-recording and gaze timestamps. Since 250 Hz gaze yields ~8 samples per video frame, samples are aggregated per frame to resample gaze uniformly to ~30 fps. Section 4.4 describes plans to replace manual $\Delta_p$ estimation with an automated estimator reporting alignment uncertainty.

***Oculomotor event classification and filtering.***

The EyeLink event parser labels each sample as belonging to a fixation, a saccade, or a missing-data interval such as a blink. This preserves explicit validity and event definitions in line with common practice (Hessels et al., 2018), but our pipeline filters strictly before sequence construction: only fixation and saccade samples are retained, blinks and other missing-data samples are discarded, and gaze coordinates outside the screen boundaries (0–1920 for $x$, 0–1080 for $y$) are removed for the region analysis. Figure 2 summarizes the full path from raw exports to the four aligned input streams.

## 2.7 The Four Input Views

Each participant–trial is represented by four aligned input views that jointly cover three categories of information about a viewing episode. The framework is viewer-centered: three of the four views describe the viewer's behavior, and only the lip-region sequence describes the stimulus, entering as context rather than content to be decoded.

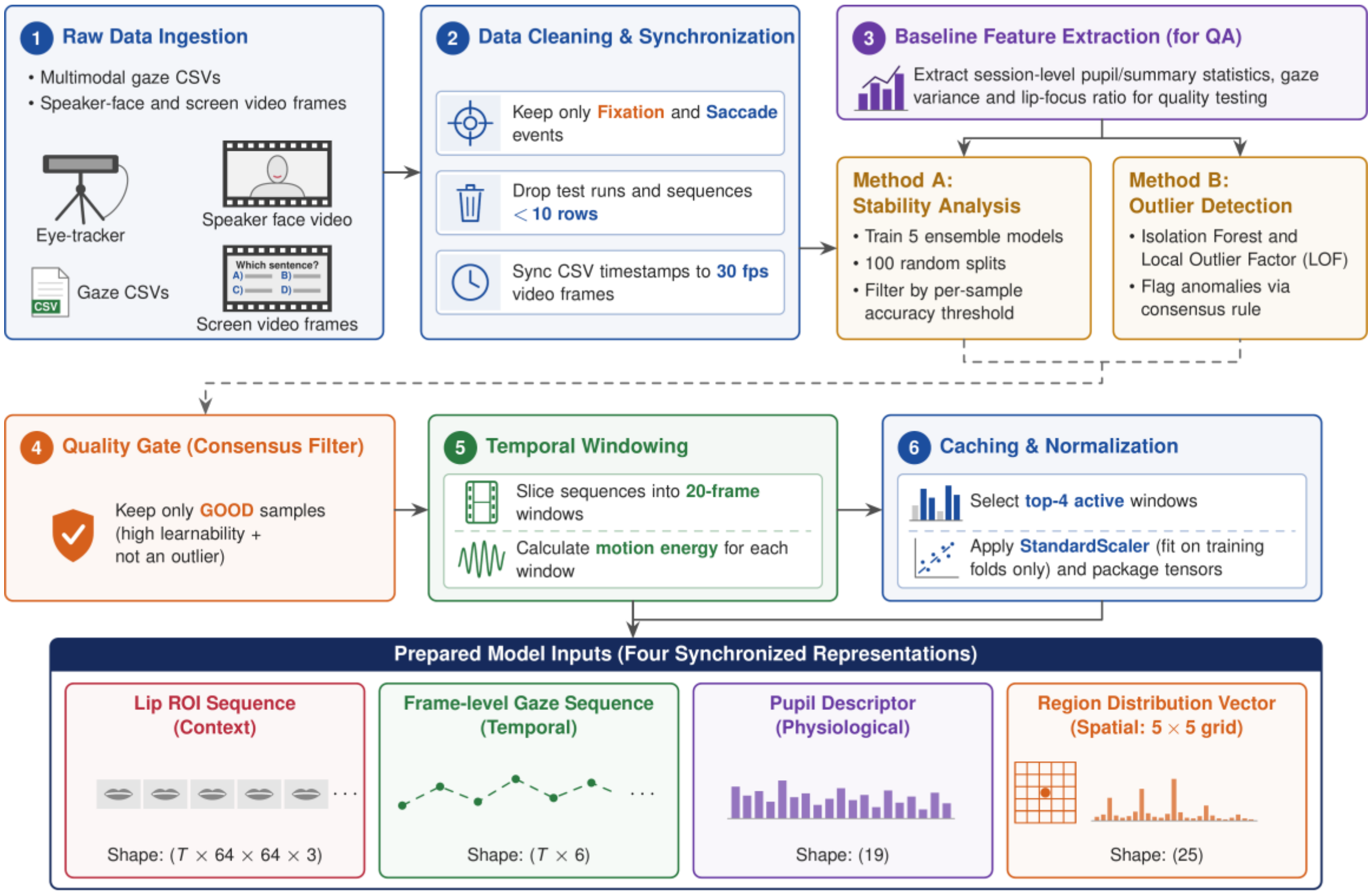

**Figure 2.** Data acquisition and preprocessing workflow. Raw eye-tracking recordings and silent lipreading videos are first cleaned: only fixation and saccade samples are retained, and test runs and sequences with fewer than 10 samples are discarded. The data then undergo synchronization, quality filtering, temporal windowing (20-frame windows, of which the four with the highest mouth-motion energy are retained per trial), and normalization (scalers fit within each training fold) to produce four synchronized input modalities: lip-region image sequences, frame-level gaze sequences, session-level pupil/summary descriptors, and session-level 5 × 5 region-distribution vectors. These representations capture complementary contextual, temporal, physiological, and spatial information for subsequent multimodal learning.

**Temporal information** is carried by the frame-level gaze sequence and the mouth motion signal derived from the lip-region crops, capturing *when* and *how* attention and articulation unfold. **Spatial information** is carried by a 5 × 5 region-proportion vector, capturing *where* attention is concentrated over the session. **Physiological information** is carried by the pupil-summary descriptor, capturing *how hard* the viewer is working (Colby & McMurray, 2021; Zekveld et al., 2018). To our knowledge, no prior study of visual speech perception models all three jointly.

Each view captures complementary information: lip-region crops encode stimulus content, gaze sequences preserve the temporal ordering of attention, pupil responses capture physiological effort, and 5

× 5 spatial distributions summarize order-free spatial allocation. In particular, two viewers can have similar spatial distributions but different scanpaths, while the spatial representation provides greater robustness to alignment errors and momentary dropout. We test the necessity of these complementary views through unimodal, bimodal, trimodal, and four-view ablations in Section 3.2. Results support the value of behavioral views, particularly gaze and pupil, while region and lip views add little or no marginal predictive value.

## 2.8 Model Architecture

The model is a four-branch multimodal architecture (Figure 3). For a participant–trial pair it receives a lip-region image sequence $I_{1:T}$, a frame-level gaze sequence $G_{1:T}$, a session-level pupil-summary vector $p$, and a session-level region-proportion vector $r$, so that a trial segment is $X = \{I_{1:T}, G_{1:T}, p, r\}$. The four branches are the Lip MotionGate (M1), the gaze encoder (M2), the pupil/summary encoder (M3), and the region-proportion encoder (M4). M1 converts the lip sequence into a mouth motion signal that gates the temporal encoding in M2, yielding a motion-gated gaze embedding $h^{(g)}$; M3 and M4 encode $p$ and $r$ into $h^{(p)}$ and $h^{(r)}$. These three embeddings reach a fusion module that combines them via learned per-trial attention weights into a shared representation $z$, from which a classifier outputs the probability of the positive class.

### *2.8.1 Lip MotionGate Branch*

For each trial we extract RGB lip-region crops from the *talker's* speaking face using fixed coordinates, resize each to 64 × 64, normalize with standard ImageNet statistics, and window them into fixed-length sequences of 20 frames; for each trial, the four windows with the highest mouth-motion energy are retained and paired with the time-aligned gaze windows (Figure 2). Restricting the input to the articulatory region excludes extraneous facial information. The branch acts purely as a two-parameter motion gate. Successive frames are converted to grayscale, the absolute frame difference $|f_t - f_{t-1}|$ captures temporal change, and a spatial mean collapses it into a frame-level motion energy scalar $e_t$. The energy is transformed into a gating signal via a learnable scale $w$ and bias $b$ so that high-motion frames pass through, low-motion frames are suppressed, and padded slots are zeroed out.

$$g_t = \sigma(w \cdot e_t + b) \tag{2}$$

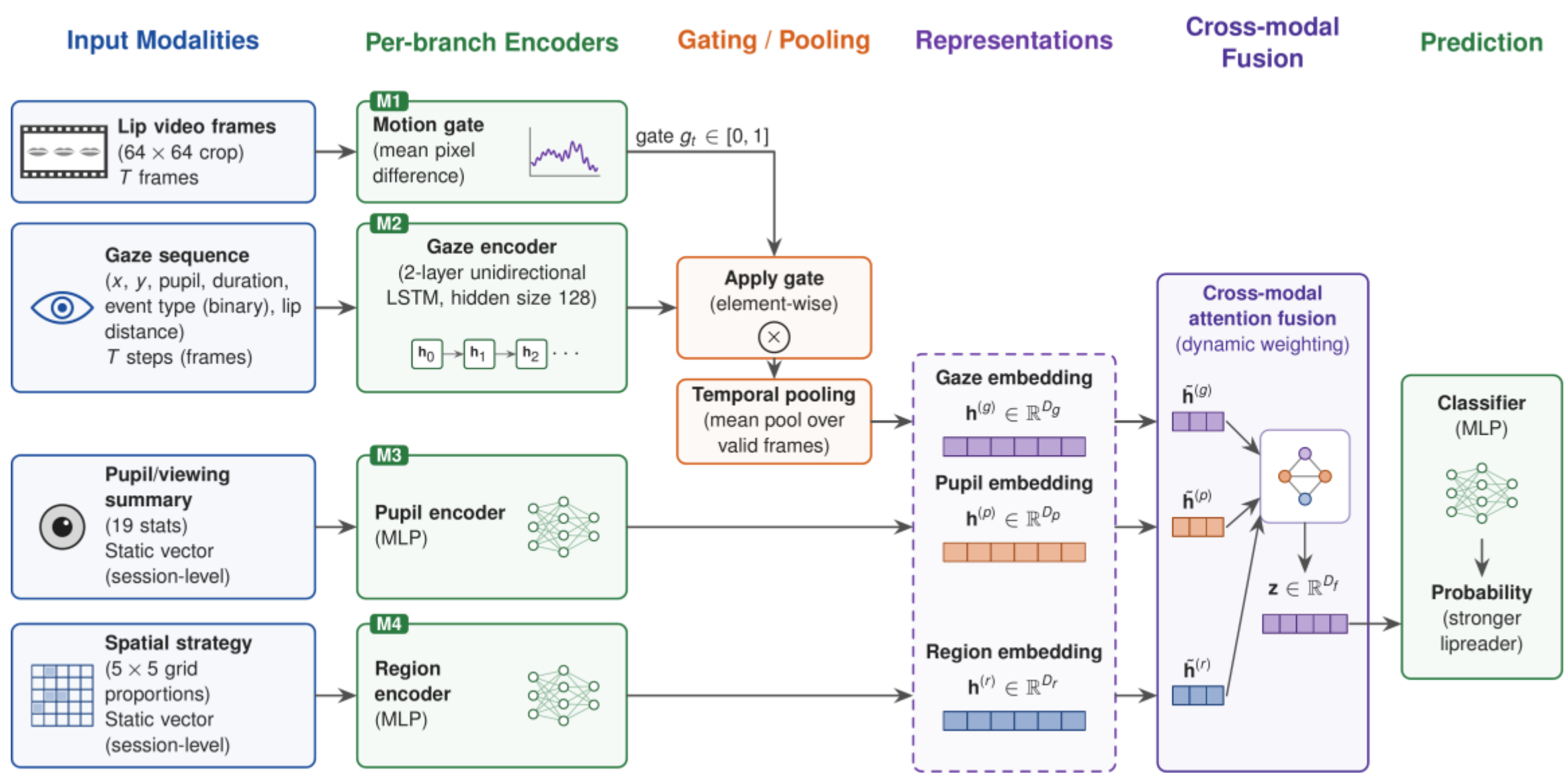

**Figure 3.** Overview of the proposed multimodal architecture. Four synchronized input streams are encoded using branch-specific networks (M1–M4). The Lip MotionGate branch (M1) generates a frame-wise motion gate $g_t$ (Equation 2) that modulates temporal gaze encoding in the gaze encoder (M2), while the pupil/summary (M3) and region-proportion (M4) branches produce complementary physiological and spatial representations. Cross-modal attention adaptively combines the three learned embeddings before binary prediction of the participant's lipreading group (stronger vs. weaker). $T$: number of time steps (frames); $D_g$: gaze embedding dimension (128); $D_p$: pupil embedding dimension (32); $D_r$: region embedding dimension (64); $D_f$: fused embedding dimension (64).

The rationale is that gaze observations are not equally informative at every frame: fixating the mouth during active articulation means something behaviorally different from fixating the same point during a pause. Gating the gaze sequence with motion energy gives high-motion intervals greater emphasis in the temporal representation. Because the stimuli depict the same talker across all participants, primitive motion energy prevents overfitting to static talker characteristics, and with two parameters the gate cannot memorize items. M1 contributes no embedding to fusion; its output modulates temporal aggregation in M2.

### *2.8.2 Gaze Encoder Module (Temporal Gaze Embedding)*

Each trial segment is a frame-aligned sequence of six gaze-related variables over a 20-frame window: gaze $x$, gaze $y$, filtered pupil diameter, gaze event duration, a strict binary eye-movement-type indicator (1 for fixation, 0 for saccade; blinks and other missing-data samples are discarded beforehand), and distance from the lip-region center. This sequence jointly captures spatial viewing behavior and event-level gaze dynamics, in line with evidence that gaze during silent visual speech is temporally structured (Benz et al., 2025; Zhong et al., 2024).

The 6-dimensional input at each timestep first passes through a projection block of linear layers, LayerNorm, ReLU, and dropout (*Linear*(6→24) → LayerNorm → ReLU → dropout → *Linear*(24→12)), reducing the dimension to 12; the projected sequence is then processed by a 2-layer unidirectional LSTM with hidden size 128. Temporal weighting comes from the external gate produced by M1: the LSTM processes all timesteps to build full temporal context, the output sequence is multiplied by the motion gate, and a masked mean-pooling operation aggregates only the real (unpadded) frames into the **Temporal Gaze Embedding** $h^{(g)} \in \mathbb{R}^{128}$. This gating logic down-weights the gaze representation during low-motion periods of speech while preserving the complete sequential context leading up to aggregation.

### *2.8.3 Pupil and Viewing-Summary Encoder Module (Pupil/Summary Embedding)*

To complement the frame-level gaze sequence, we compute a compact 19-dimensional session-level descriptor of pupil and related viewing statistics from the participant's full recording: filtered pupil diameter (mean, SD, min, max, median); left- and right-eye pupil diameter (mean, SD); pupil asymmetry (mean, SD); pupil dilation rate (mean, SD, absolute mean); gaze-to-lip-center distance (mean, SD, min, max); and a lip-focus ratio giving the proportion of gaze points on the lip region. These provide a stable session-level summary of viewing behavior (identical across a participant's segments) and are normalized with a standard scaler fit within each training fold and applied unchanged to the corresponding validation and test partitions, preventing information leakage across folds.

Because these are already session-level descriptors, M3 encodes the vector with a lightweight MLP (*Linear*(19→64)→*Linear*(64→32), ReLU, dropout), yielding the **Pupil/Summary Embedding** $h^{(p)} = f_p(p) \in \mathbb{R}^{32}$. Given the established role of pupil dilation as an index of processing effort (Colby & McMurray, 2021; Zekveld et al., 2018), M3 carries the physiological category of information identified in Section 2.7.

### *2.8.4 Region-Proportion Encoder Module (Spatial Gaze Embedding)*

We divide the 1920 × 1080 screen into a 5 × 5 grid of 25 cells of approximately 384 × 216 pixels each. Cells of this size exceed residual calibration error while still separating the regions the perceptual literature distinguishes (mouth, eyes, rest of face, background) (Viktorsson et al., 2023; Worster et al., 2018; Yi et al., 2013); the grid resolution was fixed in advance and not tuned. We compute, over the participant's full recording, the proportion of valid, on-screen gaze points falling in each cell, yielding a 25-dimensional vector summing to 1. M4 encodes this vector with an MLP of two linear layers with ReLU and dropout (*Linear*(25→64)→*Linear*(64→64)), producing the **Spatial Gaze Embedding** $h^{(r)} = f_r(r) \in \mathbb{R}^{64}$. This records the coarse spatial allocation of attention over the full recording, complementing M2 as established in Section 2.7 ($h^{(r)}$: where attention concentrated; $h^{(g)}$: how it changed over time).

### *2.8.5 Cross-Modal Attention Fusion*

Each embedding is projected to a common 64-dimensional fusion space via dedicated linear layers, yielding $\tilde{h}^{(g)}$, $\tilde{h}^{(p)}$, and $\tilde{h}^{(r)}$. The three projections are concatenated into a 192-dimensional vector and passed through an attention network (*Linear*(192→64), Tanh, *Linear*(64→3), Softmax), producing per-trial weights $w_g, w_p, w_r$. The fused representation is

$$z = w_g\tilde{h}^{(g)} + w_p\tilde{h}^{(p)} + w_r\tilde{h}^{(r)} \quad (3)$$

This lets the model emphasize the most informative modality for each trial.

### *2.8.6 Classification Head and Objective*

The fused representation $z$ passes to a lightweight classifier head: *Linear*(64→32), ReLU, dropout (0.1), *Linear*(32→1), and sigmoid, $\hat{y} = \sigma(f_c(z))$. A shallow head suits the modest labeled sample and keeps ablation effects attributable to the branches.

Training uses a Label-Smoothed Binary Cross-Entropy loss (smoothing 0.1) with a computed positive class weight. An auxiliary gate loss (weight 0.05) comprises a sparsity term penalizing deviation from a target mean gate value of 0.3 over real frames, and a smoothness term (weight 0.5) penalizing large frame-to-frame gate fluctuations, encoding the priors that only a minority of frames carry substantial articulatory motion and that articulation is temporally continuous.

### 2.9 Training and Evaluation Protocol

We evaluate the model using a participant-disjoint, 5-fold stratified cross-validation setup in which folds are formed from participants rather than segments. To ensure high-quality inputs, we use only segments that pass our data quality assessment, a consensus rule requiring samples to pass both ensemble stability thresholding (Northcutt et al., 2021; Swayamdipta et al., 2020) and dual outlier detection (Isolation Forest (Liu et al., 2008) and Local Outlier Factor (Breunig et al., 2000)), and possess valid data across active temporal windows. For each rotation, three folds are used for training, one for validation, and one for testing (approximately 60%, 20%, and 20% of participants), ensuring five non-overlapping test evaluations. We report the final metrics as the mean ± standard deviation across these 5 rotations consistent with current guidance on validating machine learning models under small samples in speech, language, and hearing research (Ghasemzadeh et al., 2024).

We optimize using Adam (Kingma & Ba, 2014) (learning rate $10^{-3}$, weight decay $10^{-5}$) with a StepLR scheduler (halving the learning rate every 15 epochs) and gradient clipping (norm = 1.0). Training runs for up to 100 epochs with early stopping (patience = 20) based on validation Balanced Accuracy. For each fold, the optimal classification threshold is selected on the validation set by sweeping values from 0.1 to 0.85 (step = 0.05) and then applied to the test set. The classification task uses the extreme-groups subset (8 stronger and 10 weaker lipreaders; N = 18), contributing 360 participant–clip segments (18 participants × 20 clips, each represented by its four most active 20-frame windows), all of which passed the quality filter. Because segments from the same participant are correlated, all cross-validation splits are participant-disjoint (not segment-disjoint), and reported accuracy, F1, and AUC should be interpreted as estimates of participant-level discriminability from an effective labeled sample of 18 participants, not 360 independent observations.

### 2.10 Model Configurations Evaluated

We evaluate the proposed multimodal formulation (**Full model / attention**, Table 2) alongside ablation variants spanning the unimodal, bimodal, and fusion-strategy configurations needed to test the complementarity claim of Section 2.7; each configuration below is named identically to its row in Table 2.

- **Unimodal baselines** (*Gaze only*, *Pupil/summary only*, *Region only*): independent evaluation of the motion-gated gaze sequence via the unidirectional LSTM (this model retains the lip-derived motion gate), the pupil/summary vector via MLP, and the region proportions via MLP.
- **Reduced multimodal** (*No gaze*, *No pupil/summary*, *No region*, *No lip (no gate)*): the full architecture with one branch removed, or with the Lip MotionGate disabled to test unweighted gaze-sequence temporal pooling.
- **Fusion variants** (*Concatenation*, *Average pooling*, *Early fusion*, vs. the proposed *Full model / attention*): dynamic attention fusion compared against fixed integration rules.

### 2.11 Evaluation Metrics and Statistical Analysis

We report Accuracy, F1 Score, and Area Under the ROC Curve (AUC) computed on the held-out test folds, aggregating the mean and standard deviation across the 5 cross-validation rotations. Because standard accuracy can be misleading under class imbalance, we use validation Balanced Accuracy at the default 0.5 threshold for early stopping and select the operating threshold that maximizes validation Balanced Accuracy in the sweep described in Section 2.9. The F1 Score and AUC provide complementary insights: AUC summarizes discriminative ability across all classification thresholds, although under class imbalance precision–recall-based measures such as F1 can be more informative (Saito & Rehmsmeier, 2015).

Classification metrics are descriptive summaries across cross-validation folds rather than inferential tests. Because the labeled set contains 10 weaker and 8 stronger lipreaders, we take the majority-class rate ($10/18 \approx 0.56$) as the chance reference for accuracy and 0.50 for AUC; "above chance" therefore denotes exceeding these reference values, not a significance test. Feature attributions (Section 3.3) were computed with Integrated Gradients (Sundararajan et al., 2017) and averaged across segments; positive values push the prediction toward the stronger-lipreader class (Class 1) and negative values toward the weaker-lipreader class (Class 0). Group differences in viewing behavior are reported descriptively, with Cohen's d as the effect size.

## 3. Results

The experiments below first establish that lipreading skill is predictable from viewing behavior and that learned fusion outperforms fixed rules in accuracy (Section 3.1), then identify which signals carry the information (Section 3.2), examine whether the model relies on behaviorally plausible cues (Section 3.3), and summarize the ablation findings (Section 3.4).

### 3.1 Overall Performance and Fusion

Table 2 reports the full model and its ablations. The four-branch attention-fusion model attains accuracy $0.680 \pm 0.040$, F1 $0.547 \pm 0.022$, and AUC $0.700 \pm 0.068$, above the chance reference values (Section 2.11). As noted in Section 2.9, these values estimate participant-level discriminability from 18 labeled participants, but they are sufficient to motivate the remaining analyses. In accuracy, learned attention fusion outperforms naive alternatives: concatenation reaches only 0.581 accuracy and early fusion 0.600, while average pooling (0.658) trails the learned weighting (0.680). In AUC, however, the fusion variants are comparable (0.675–0.709), with concatenation slightly above the full model. The large SD of the concatenation baseline ($\pm 0.132$) indicates that naive fusion is also less stable across folds. The fused model is competitive with, but does not dramatically exceed, the best single-modality baseline, indicating that the branches contribute unequally, which motivates the analysis below.

**Table 2.** Ablation of model configurations across modalities. Best value per metric in bold; ↑ indicates higher is better. Values are mean ± SD, computed from participant-disjoint 5-fold cross-validation (N = 18 labeled participants; see Section 2.9).

| Model configuration | Modality | Accuracy ↑ | F1 ↑ | AUC ↑ |
|---|---|---|---|---|
| Full model / attention (proposed) | Multi | **0.680 ± 0.040** | 0.547 ± 0.022 | 0.700 ± 0.068 |
| No lip (no gate) | Multi | 0.674 ± 0.050 | **0.562 ± 0.047** | **0.711 ± 0.063** |
| Concatenation | Multi | 0.581 ± 0.132 | 0.540 ± 0.058 | 0.709 ± 0.077 |
| Average pooling | Multi | 0.658 ± 0.062 | 0.501 ± 0.098 | 0.689 ± 0.074 |
| Early fusion | Multi | 0.600 ± 0.073 | 0.492 ± 0.087 | 0.675 ± 0.076 |
| No gaze (pupil + region) | Bi | 0.656 ± 0.057 | **0.562 ± 0.017** | 0.706 ± 0.063 |
| No pupil/summary | Bi | 0.618 ± 0.066 | 0.428 ± 0.118 | 0.623 ± 0.059 |
| No region | Bi | 0.665 ± 0.039 | 0.504 ± 0.082 | 0.702 ± 0.065 |
| Gaze only | Uni | 0.643 ± 0.054 | 0.465 ± 0.082 | 0.643 ± 0.083 |
| Pupil/summary only | Uni | 0.675 ± 0.024 | 0.496 ± 0.127 | 0.680 ± 0.074 |
| Region only | Uni | 0.581 ± 0.077 | 0.463 ± 0.086 | 0.602 ± 0.029 |

### 3.2 Which Signals Carry the Information?

Removing the pupil/summary branch causes the largest degradation: accuracy drops from 0.680 to 0.618, F1 from 0.547 to 0.428, and AUC from 0.700 to 0.623. Correspondingly, the pupil/summary-only model is the strongest unimodal baseline (0.675 accuracy, 0.680 AUC) and remarkably stable across folds (SD 0.024). The pupil-and-viewing-summary branch is therefore the most informative of the three behavioral branches. This branch is not purely pupillary: as described in Section 2.8.3, four of its 19 descriptors summarize gaze-to-lip-center distance and one is the lip-focus ratio, so the ablation isolates the branch rather than pupil size alone. The gaze sequence contributes moderately alone (0.643 accuracy) and adds accuracy in combination (0.680 vs. 0.656 without it, although F1 and AUC are slightly higher without it), while region proportions are the weakest branch (0.581 accuracy) and add little at the margin (accuracy drops from 0.680 to 0.665 when removed).

These conclusions are supported by two independent lines of evidence: the leave-one-out rows show what is lost when a signal is removed, and the unimodal rows show what can be achieved with that signal alone. For the pupil/summary branch both agree, confirming that it carries information not captured by the gaze-sequence or region views. The complementarity claim holds for the behavioral views, where learned fusion achieves higher accuracy than every single view and every fixed rule (though not the highest F1 or AUC), but is qualified for the region view, which adds little.

The lip-image stream presents an instructive negative result. Removing it barely changes accuracy (0.674) and slightly helps on F1 (0.562) and AUC (0.711). Because the lip crops depict the talker and are identical across participants for a given stimulus, they encode item difficulty but cannot separate strong from weak lipreaders. The participant-discriminative signal resides in the behavioral branches.

### 3.3 Interpretability

We examine whether the model relies on behaviorally plausible cues through two complementary types of interpretability analysis: **(1) data/stimulus-level interpretability**, heatmap visualizations of articulatory motion (Figure 4), and **(2) feature-level interpretability**, attribution analysis of group-level differences in gaze and pupil features (Figure 5). We use Integrated Gradients (Section 2.11) together with group comparisons of viewing behavior to identify the temporal, spatial, and physiological cues driving predictions. Figure 4 shows frame-level motion-energy heatmaps used by the Lip MotionGate (Section 2.8.1) across four representative intervals: baseline articulation, high motion, a stressed syllable, and lower motion. Each panel shows two consecutive frames and their absolute difference image, which feeds the gate's motion-energy scalar (Equation 2); brighter regions indicate greater local motion, particularly around the lip contour. Read sequentially, the panels visualize the mouth-motion signal used by the gate. Together with the quantitative ablations in Table 2, these visualizations help explain why pupil- and gaze-linked branches, rather than raw lip appearance, carry participant-discriminative information: motion energy tracks stimulus-driven articulation, while gaze and pupil behavior track the viewer's response.

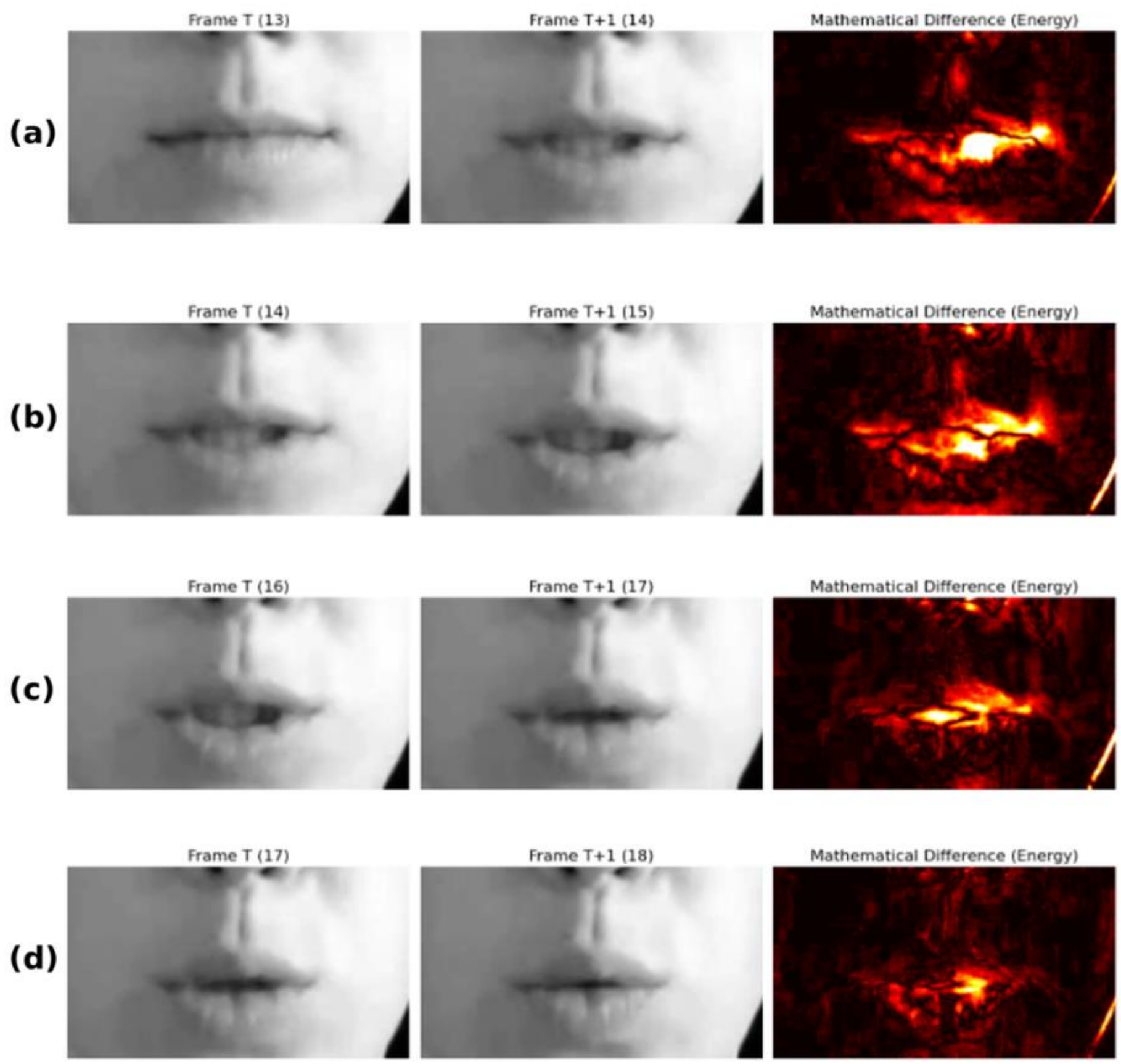

**Figure 4.** Frame-level motion-energy heatmaps used by the Lip MotionGate (M1) for four representative frame pairs: (a) baseline articulation, with motion concentrated around the lip aperture; (b) a high-motion articulatory interval, with the difference image bounded within the lip aperture; (c) a stressed syllable, with a sharp motion peak at the lip contour; and (d) a lower-motion interval, with a weaker, more diffuse difference image. Panels (a)–(d) run top to bottom; each shows two consecutive frames (titled "Frame T" and "Frame T+1," with their frame indices) and their absolute difference image (titled "Mathematical Difference (Energy)"), which feeds the gate's motion-energy scalar (Equation 2). The panels show the talker's mouth only; viewer gaze is not depicted.

Counterintuitively, the aggregate time spent looking at the general mouth area does not differ between the groups: participants in the stronger (Class 1) and weaker (Class 0) groups spent nearly identical proportions of their trial time looking at the lower face (73.17% vs. 73.21%, Cohen's $d = 0.002$), indicating that broad lip-directed gaze is common to both groups but insufficient to distinguish stronger from weaker lipreaders. The pupil/summary branch instead carries the most informative behavioral signal in our data. Within this branch, Integrated Gradients assign the largest positive attributions (toward Class 1) to mean pupil diameter (left, right, and combined). Conversely, the gaze-to-lip distance descriptors in the same branch, alongside the standard deviation of pupil asymmetry, carry negative attributions (toward Class 0), possibly reflecting differences in cognitive engagement or processing load.

Spatially, the attributions are concentrated in a few cells of the 5 × 5 grid. Gaze in the cell containing the core articulator region (R3C2) carries the largest positive attribution within the spatial branch, whereas gaze in the central mid-face cell (R2C2) carries the largest negative attribution. These should be read as model-based feature associations rather than as evidence that stronger lipreaders maintain a more precise fixation strategy, for two reasons. First, the coarse spatial branch is the weakest of the three behavioral branches in the ablation analysis (Table 2). Second, the attribution patterns are not straightforwardly consistent: within the temporal gaze branch, distance from the lip center carried positive attribution, whereas within the spatial branch a single articulator-region cell contributed positively. We therefore do not advance spatial precision as a conclusion of this study.

Figure 5 presents the mean attribution scores for the three behavioral branches. Positive values indicate features that increase the likelihood of predicting a stronger lipreader (Class 1); negative values contribute toward a weaker lipreader (Class 0).

In the temporal gaze branch (Figure 5a), distance to the lip center carries positive attribution, while filtered pupil diameter and gaze event duration carry negative attribution. In the pupil/summary branch (Figure 5b), mean pupil diameter (left, right, and combined) carries the largest positive attributions, while the gaze-to-lip distance descriptors and pupil-asymmetry variability carry negative attributions. In the spatial branch (Figure 5c), only a few cells contribute: gaze in the core articulator cell (R3C2) contributes toward the stronger-lipreader class, and gaze in the central mid-face cell (R2C2) toward the weaker-lipreader class.

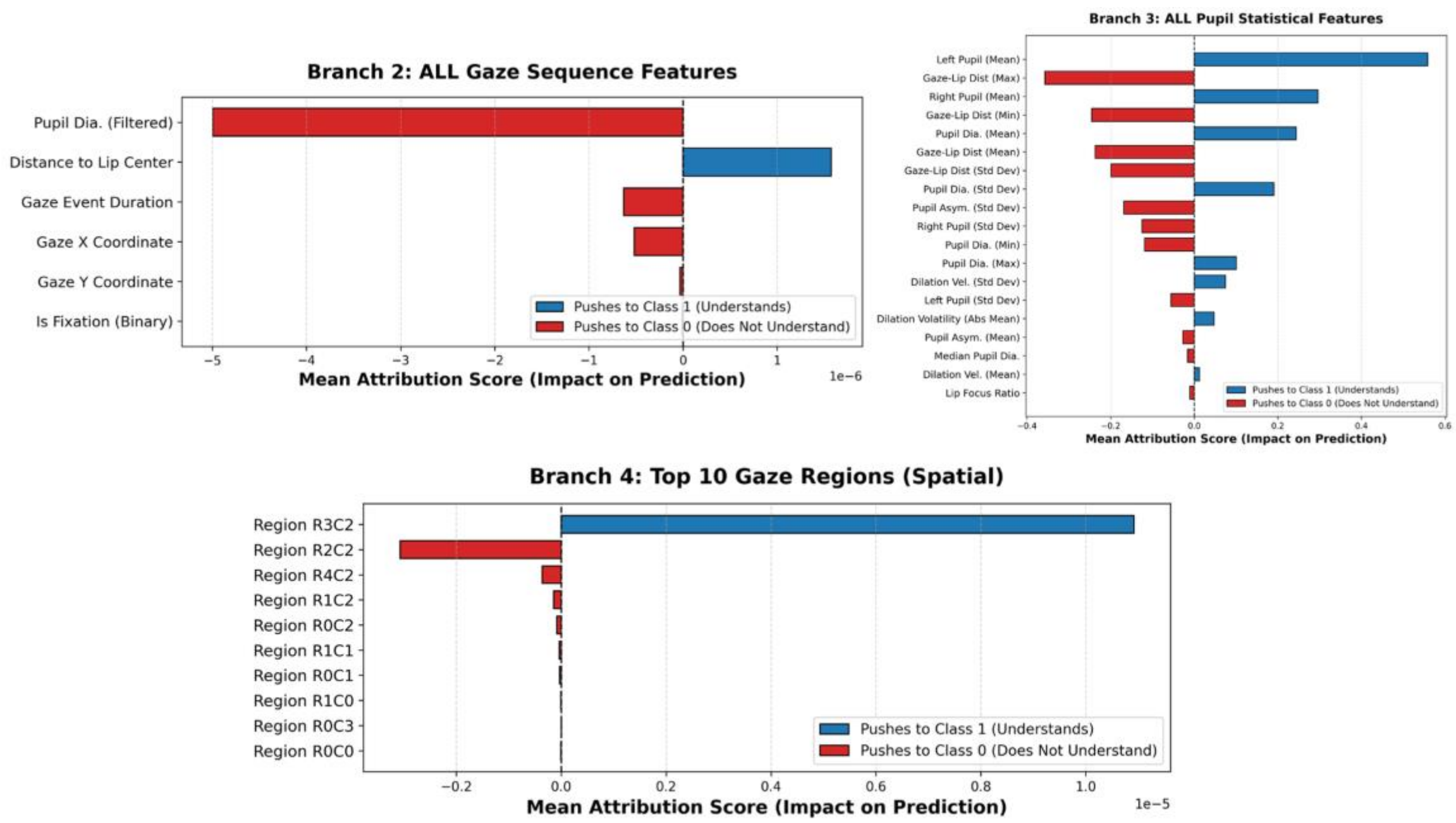

**Figure 5.** Feature-based behavioral interpretability using attribution analysis. Mean attribution scores are shown for the three behavioral branches. Positive values (blue) indicate features contributing toward Class 1 (stronger lipreaders), and negative values (red) indicate features contributing toward Class 0 (weaker lipreaders). (a) Temporal gaze branch (M2; "Branch 2," top left), (b) pupil/summary branch (M3; "Branch 3," top right), and (c) coarse 5 × 5 spatial branch (M4; "Branch 4," bottom; the 10 top-ranked cells are shown, indexed by row R and column C from the top-left cell, R0C0). In the legends, "Understands" and "Does Not Understand" denote the stronger- and weaker-lipreader classes. Attribution scales differ across panels (note the $\times 10^{-6}$ and $\times 10^{-5}$ axis multipliers in a and c).

### 3.4 Summary of Ablation Findings

Taken together, the ablations reported in Table 2 show that lipreading skill is predictable above chance from viewing behavior, with attention-based fusion achieving higher accuracy than fixed fusion strategies. Predictive information is driven primarily by the pupil-and-viewing-summary branch, with the frame-level gaze sequence adding accuracy, whereas coarse spatial allocation contributes less. The lip-image stream provides limited participant-level information because the stimulus is shared across participants. The gaze sequence view also outperformed the aggregate spatial summary. Because these two views differ in feature content as well as in temporal structure, this comparison indicates that the sequential view is more informative but does not isolate the contribution of temporal ordering.

## 4. Discussion

### 4.1 What the Results Say About Human Lipreading

Three findings address our central question: why do some individuals extract speech from vision alone while others watching the same face do not? First, stronger and weaker lipreaders differ in viewing behavior, distinguishable above chance from a short viewing episode. Second, these differences appear to reflect effort, indexed by pupil-linked signals, more than aim: pupil summaries were most informative, while coarse spatial viewing was least informative. Third, time-resolved gaze outperformed aggregate spatial summaries, suggesting a role for coordination between attention and articulation, although temporal ordering cannot be isolated from feature content (Section 3.4). Thus, lipreading may involve sustained, temporally coordinated effort rather than simply targeting informative regions. This may explain why mouth-dwell proportion correlates with speechreading in some populations (Worster et al., 2018) but not others (Lansing & McConkie, 2003; Wilson et al., 2016), as dwell proportion may be a coarse proxy for dynamic, effortful processes.

***The pupil and viewing-summary finding.***

Pupil dilation is an established index of listening effort (Colby & McMurray, 2021; Kuchinsky et al., 2023; Zekveld et al., 2018) and has supported hearing-status classification (Lebiecka-Johansen et al., 2025). The pupil/summary branch was individually strongest, and its removal most damaging. In our silent paradigm, pupil-linked variation may therefore reflect individual differences in visual speech processing. However, the branch combines pupil and gaze features, and pupil size is luminance-sensitive. Mean pupil diameter had the strongest positive attributions toward stronger lipreaders, while gaze-to-lip distance and pupil-asymmetry variability had negative attributions.

***The gaze finding.***

Frame-level gaze outperformed aggregate region proportions, indicating discriminative information beyond dwell time, although temporal ordering cannot be isolated from feature content. This aligns with evidence that eye movements during silent speech track latent acoustic structure (Benz et al., 2025) and gaze relates to cortical tracking of audiovisual speech (Tan et al., 2022). It may explain inconsistent associations between mouth-dwell proportion and speechreading (Wilson et al., 2016): predictive information may lie in how gaze unfolds, not just where it lands.

***The lip-image null result.***

Lip crops are shared across participants for each stimulus, encoding item difficulty but not participant-specific information. Removing the lip branch left accuracy essentially unchanged (0.674 vs. 0.680), with slightly improved F1 and AUC. This distinguishes shared stimulus difficulty from participant-specific behavior, the latter being more discriminative for behavioral phenotyping. Future work could incorporate visual-stream information about item difficulty or model the participant's own face. The talker's face remains useful as timing information, as used by the MotionGate.

## 4.2 Qualitative Debrief

After each session, participants informally described their lipreading strategy. Most reported focusing on a single sentence position or on articulatory cues such as mouth opening and lip movements. Several described the task as effortful. None mentioned the pupil-linked or gaze-to-lip-distance features that Integrated Gradients associated with stronger versus weaker lipreaders, suggesting that objective gaze- and pupil-based measures can complement self-report.

## 4.3 Fusion, Interpretability, and Translational Implications

Beyond accuracy, branch ablations and feature attributions provide a transparent, auditable account of the signals used by the model (Altinok, 2024), with potential applications in clinical and assistive settings. Since speechreading improves with training (Bernstein, Auer, & Eberhardt, 2022; Buchanan-Worster et al., 2021), identifying stronger lipreaders' viewing behaviors could inform gaze-aware training feedback, adaptive assistive technologies, and clinical tools supplementing self-report with brief silent-viewing tasks. Our findings also suggest that "look at the mouth" instructions and dwell-proportion measures may be less sensitive than pupil-linked and frame-level gaze measures. However, the observational design supports prediction, not causality; demonstrating benefits from gaze-aware feedback requires an intervention study.

As participants were recruited from the general adult population without hearing assessment, these applications remain hypotheses to be tested in the project's planned hard-of-hearing populations.

### 4.4 Limitations

The labeled set is modest and drawn from a single site, so the results establish feasibility rather than population-level generalization; we mitigate leakage with participant-disjoint cross-validation, but larger multi-site data are needed. Recruitment nonetheless spanned nine countries with within-group linguistic diversity. The extreme-groups labeling sharpens the contrast but discards mid-range participants and may inflate apparent separability, making graded regression an important next step. Frame-level alignment relies on a manually estimated offset $\Delta_p$, which future work will automate. Cross-device generalization is not yet evaluated; a Tobii extension will test whether pupil-linked effort remains stable across acquisition settings. Pupil measures are sensitive to luminance, so we treat pupil features as effort-related under controlled luminance rather than as pure cognitive indices. Two features of the analysis pipeline also qualify the results. The pupil-summary branch contains four gaze-to-lip-distance descriptors and the lip-focus ratio in addition to the pupil measures (Section 2.8.3), so its ablation tests the composite branch rather than pupil size in isolation. Because the pupil/summary and region vectors are computed per session, these branches cannot capture trial-to-trial variation within a participant. Hearing status was not formally assessed. Finally, our stimuli are visual-only; extending to audiovisual stimuli and confirmed hard-of-hearing samples would strengthen the clinical interpretation.

### 4.5 Ethics and Data Governance

The LiPS dataset contains sensitive biometric signals (facial video, gaze trajectories, and pupil size). Access and sharing are governed by a data-sharing agreement, and any release will prioritize privacy by sharing derived features and anonymized summaries rather than raw identifiable recordings. The study protocol was approved by the Regional Medical Ethical Board of the Northern Ostrobothnia Hospital District, Finland (ethics board reference no. 47/2022). The study was prospectively registered with ClinicalTrials.gov (NCT05854719).

**4.6 Conclusions**

This work demonstrates that individual differences in lipreading ability are reflected in measurable viewing behavior during silent visual speech and can be predicted above chance from synchronized eye-tracking signals. The proposed multimodal framework achieves $0.680 \pm 0.040$ accuracy through adaptive attention fusion of temporal, spatial, and physiological representations of viewing behavior, closely matched by the pupil/summary-only baseline (0.675), indicating that fusion adds robustness and interpretability more than raw predictive gain. Systematic ablation and interpretability analyses indicate that the pupil-and-viewing-summary branch and the frame-level gaze sequence carry most of the discriminative information, while the talker's mouth appearance contributes little to participant-level discrimination. These findings provide quantitative evidence that successful visual speech perception is characterized by distinctive behavioral strategies rather than stimulus appearance alone, strengthening the connection between multimodal machine learning and speech-hearing science. Future work will extend the framework through automated gaze–video synchronization, cross-device evaluation, and continuous measures of lipreading ability, with the broader goal of providing objective behavioral evidence to support assessment and training for people who are deaf or hard of hearing.

## Acknowledgments

This work was a collaborative effort among Center for Machine Vision and Signal Analysis (CMVS), University of Oulu, Finland, Indian Institute of Technology Ropar, India and Zhejiang University, China. This work was in part financially supported by the Eudaimonia Institute of the University of Oulu. Grammarly has been used to support language editing. No AI-generated text, images or figures are included.

## Conflict of Interest

The authors declare no conflicts of interest.

## Data Availability Statement

The LiPS dataset contains sensitive biometric data (facial video, eye-tracking recordings, and pupil measurements). Access is restricted and governed by data-sharing agreements. Derived features and anonymized summaries may be made available upon reasonable request to the corresponding author, subject to ethical approval and institutional data-governance requirements.